\documentclass[conference]{IEEEtran}
\IEEEoverridecommandlockouts

\usepackage{cite}
\usepackage{amsmath,amssymb,amsfonts}
\usepackage{algorithmic}
\usepackage{graphicx}
\usepackage{textcomp}
\usepackage{xcolor}
\usepackage{hyperref}
\def\BibTeX{{\rm B\kern-.05em{\sc i\kern-.025em b}\kern-.08em
    T\kern-.1667em\lower.7ex\hbox{E}\kern-.125emX}}
\begin{document}

\title{D-GATNet: Interpretable Temporal Graph Attention Learning for ADHD Identification Using Dynamic Functional Connectivity\\

\thanks{Identify applicable funding agency here. If none, delete this.}
}

\author{
\IEEEauthorblockN{
Qurat Ul Ain\textsuperscript{1,2},
Alptekin Temizel\textsuperscript{2},
Soyiba Jawed\textsuperscript{1}
}

\IEEEauthorblockA{
\textsuperscript{1}Department of Computer and Software Engineering,
College of Electrical and Mechanical Engineering,\\
National University of Sciences and Technology,
Islamabad 44000, Pakistan
}

\IEEEauthorblockA{
\textsuperscript{2}Graduate School of Informatics,
Middle East Technical University, 06800, Ankara, Turkey
}

\IEEEauthorblockA{
qain.cse22ceme@student.nust.edu.pk,
atemizel@metu.edu.tr,
soyiba.jawed@ceme.nust.edu.pk
}
}

\maketitle

\begin{abstract}
Attention Deficit Hyperactivity Disorder (ADHD) is a prevalent neurodevelopmental disorder whose neuroimaging-based diagnosis remains challenging due to complex time-varying disruptions in brain connectivity. Functional MRI (fMRI) provides a powerful non-invasive modality for identifying functional alterations. Existing deep learning (DL) studies employ diverse neuroimaging features; however, static functional connectivity remains widely used, whereas dynamic connectivity modeling is comparatively underexplored. Moreover, many DL models lack interpretability. In this work, we propose D-GATNet, an interpretable temporal graph-based framework for automated ADHD classification using dynamic functional connectivity (dFC). Sliding-window Pearson correlation constructs sequences of functional brain graphs with regions of interest as nodes and connectivity strengths as edges. Spatial dependencies are learned via a multi-layer Graph Attention Network, while temporal dynamics are modeled using 1D convolution followed by temporal attention. Interpretability is achieved through graph attention weights revealing dominant ROI interactions, ROI importance scores identifying influential regions, and temporal attention emphasizing informative connectivity segments. Experiments on the Peking University site of the ADHD-200 dataset using stratified 10-fold cross-validation with a 5-seed ensemble achieved 85.18\% ± 5.64 balanced accuracy and 0.881 AUC, outperforming state-of-the-art methods. Attention analysis reveals cerebellar and default mode network disruptions, indicating potential neuroimaging biomarkers.

\end{abstract}


\section{Introduction}
Attention Deficit Hyperactivity Disorder (ADHD) is a prevalent neurodevelopmental disorder in children, characterized by inattention and hyperactivity~\cite{b1}. Early diagnosis is crucial for reducing long-term adverse outcomes~\cite{b2}. Recent advances in neuroimaging-based research, driven by machine learning (ML) and deep learning (DL), have improved automated ADHD   classification~\cite{b3,b4}. Among neuroimaging modalities, functional magnetic resonance imaging (fMRI) has emerged as a promising auxiliary diagnostic tool. As a non-invasive technique with high spatial resolution and acceptable temporal resolution, fMRI is widely used to detect structural and functional brain alterations~\cite{b5}~\cite{b6}. Consequently, DL frameworks have been increasingly applied to fMRI data to enhance ADHD diagnosis. For example, Zhao et al.~\cite{b7} proposed a dynamic graph convolutional network (dgCN) based on functional connectivity (FC) brain graphs, evaluated using 10-fold cross-validation (CV), and also interpreted potential ADHD biomarkers. Qiang et al.~\cite{b8} introduced a Spatiotemporal Attention Autoencoder (STAAE) using resting-state temporal templates evaluated under 5 fold CV. Chen et al.~\cite{b9} developed Att-AENet for FC based biomarker learning using leave-one-out-cross-validation (LOOCV) with interpretable region identification. Peng et al.~\cite{b10} proposed Summation-based Synergic Artificial Neural Network  (SSANN) with dual 3D-CNN branches for the fusion of sMRI and rs-fMRI data, evaluated using an 80:20 hold-out split. Wang et al.~\cite{b11} proposed two approaches, ICA-CNN and correlation autoencoder, evaluated under hold out validation. Pei et al.~\cite{b6} introduced Gaussian noise, Mix-up, and sliding-window augmentation and developed a CNN–GAT fusion model leveraging FC features, with the sliding window strategy demonstrating superior performance among the augmentation methods. Zhang et al.~\cite{b12} proposed a Diffusion Kernel Attention Transformer framework, evaluated using an 80:20 data split, and performed interpretability analysis. Zhang et al.~\cite{b13} proposed a model named BNC-DGHL, which incorporates CNN and hashing layers and evaluated under 5-fold CV. Hu et al.~\cite{b14} introduced Source Free Semi-Supervised Transfer Learning (S3TL) with LSTM-based region of interest (ROI) features, evaluated using 10 fold CV. Gulhan and Özmen~\cite{b15} applied 3D-CNNs to faLFF/ReHo measures evaluated with 5 fold CV.

\par
Despite this progress, robust ADHD diagnosis remains challenging. Many studies reporting strong performance rely on LOOCV ~\cite{b9}, which is computationally expensive and may produce high variance estimates that limit generalization. Moreover, although diverse features have been explored, static functional connectivity (FC) remains the most widely adopted representation ~\cite{b16}. However, static FC ignores the temporal dynamics of rs-fMRI, even though ADHD related abnormalities may emerge through time varying interactions. Although dynamic functional connectivity (dFC) has been examined in a few recent works ~\cite{b17},  its application to automated ADHD diagnosis remains comparatively underexplored, despite showing superiority over static measures in other psychiatric disorders ~\cite{b18}. In addition, most DL models lack interpretability, providing limited insight into the spatio-temporal patterns driving predictions. Therefore, we propose an interpretable resting-state functional MRI (rs-fMRI) driven ADHD classification framework leveraging dFC to capture both spatial and temporal abnormalities in functional brain networks. The main contributions are as follows:

\begin{itemize}
    \item We develop a temporal graph-based deep learning pipeline for ADHD identification using dFC representations extracted from rs-fMRI.
    \item We model spatial interactions among brain regions through a Graph Attention Network (GAT) to learn discriminative ROI-to-ROI connectivity patterns.
    \item We incorporate temporal convolution with an attention mechanism to capture time varying connectivity dynamics and highlight informative temporal windows.
    \item We enhance clinical interpretability by exploiting graph and temporal attention weights to identify dominant ROI connections and dynamic brain states associated with ADHD.
\end{itemize}

\section{Proposed Methodology}
\begin{figure*}[t]
    \centering
    \includegraphics[width=0.75\linewidth]{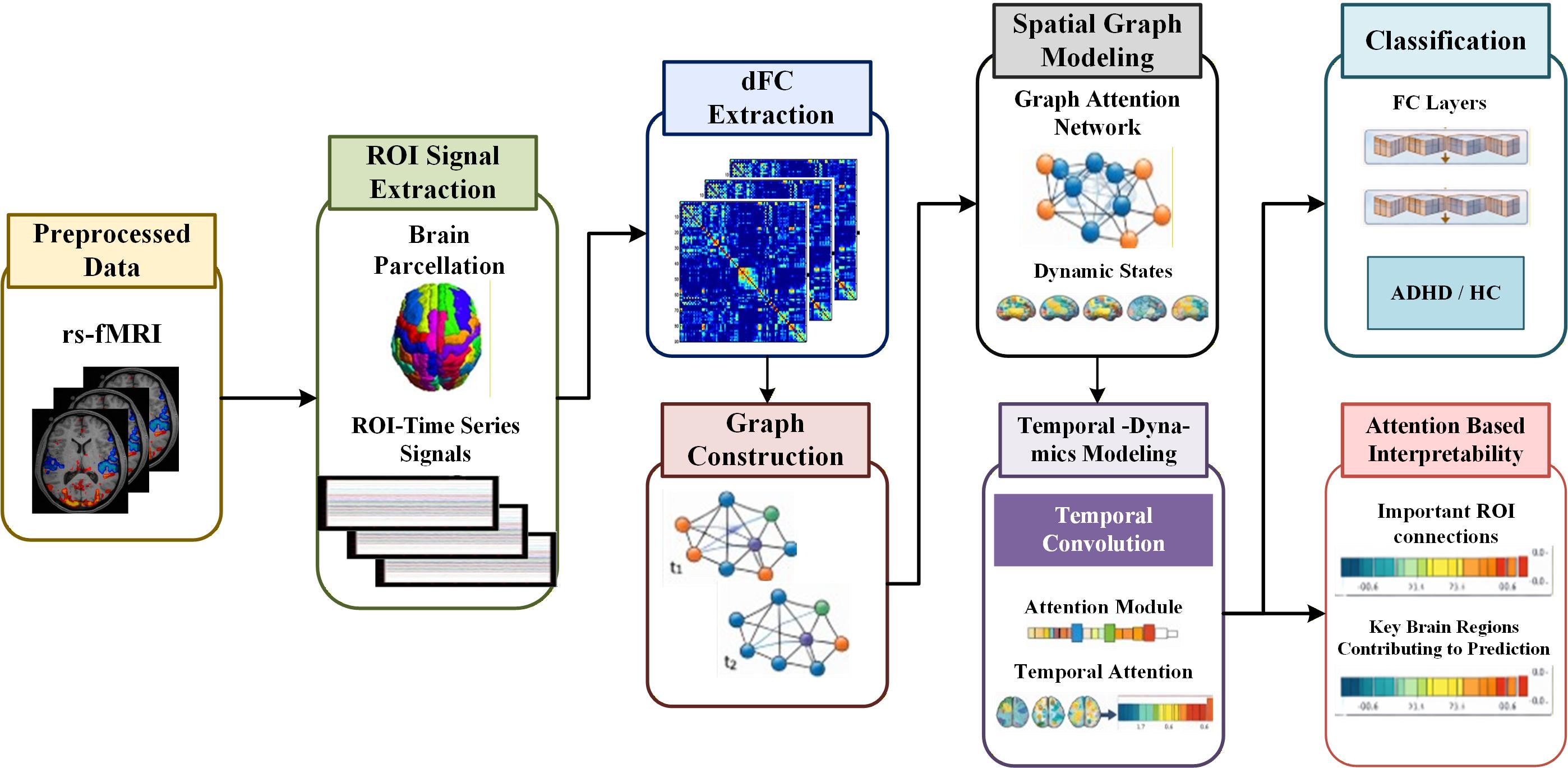}
    \caption{Overview of the proposed D-GATNet framework for ADHD classification.}
    \label{fig:overview_framework}
\end{figure*}

In this study, we propose a D-GATNet (Dynamic Graph Attention Network)  framework for automated ADHD identification using resting-state fMRI. Following preprocessing, dynamic functional connectivity (dFC) is extracted to model time varying brain connectivity patterns and capture complementary spatial and temporal information.The proposed architecture consists of five main modules: (i) Dynamic Connectivity Representation, (ii) Graph Construction, (iii) Spatial Graph Modeling, (iv) Temporal Dynamics Modeling, and (v) Classification. In addition, intrinsic interpretability is achieved through the learned graph and temporal attention weights, which highlight the most influential ROI connections and dynamic windows contributing to the final prediction.
 The overall framework is illustrated in.
 fig.~\ref{fig:overview_framework}.
\par

\subsection{Dynamic Connectivity Representation}\label{AA}
For each subject, the preprocessed rs-fMRI data are represented by an ROI-wise mean time series matrix $\mathbf{X} \in \mathbb{R}^{L \times N}$.where $N=116$ denotes the number of ROIs and $L$ is the number of time points.To characterize time-varying functional interactions between brain regions, we compute dFC using Pearson correlation within a sliding window framework. Specifically, a temporal window of size $w=40$ with step size $s=20$ is applied to the ROI time series. For each window $t$, a connectivity matrix is obtained as defined in Eq.~\ref{eq:roi_mean}.

\begin{equation}
\mathbf{F}^{(t)} = \mathrm{corr}(\mathbf{X}_{t:t+w})
\label{eq:roi_mean}
\end{equation}

where $\mathbf{F}^{(t)} \in \mathbb{R}^{N \times N}$ denotes the functional connectivity graph at window $t$.

Thus, each subject is represented as a sequence of dynamic FC matrices, as expressed in Eq.~\ref{eq:roi}.

\begin{equation}
\{\mathbf{F}^{(t)}\}_{t=1}^{T}
\label{eq:roi}
\end{equation}

where $T$ is the total number of temporal windows.


\subsection{Graph Construction}\label{AA}

Each FC matrix is interpreted as a brain graph, where ROIs correspond to nodes and connectivity strengths represent edges. Since correlation matrices are densely connected, thresholding is applied to retain the most informative connections and suppress noise. The adjacency matrix is defined according to Eq.~\ref{eq:ro}.

\begin{equation}
\mathbf{A}^{(t)}_{ij} =
\begin{cases}
1, & |\mathbf{F}^{(t)}_{ij}| \ge \tau\\
0, & \text{otherwise},
\end{cases}
\label{eq:ro}
\end{equation}

where $\tau$ preserves the top 30\% strongest functional links. Self loops are included to maintain node identity, producing a sparse graph representation suitable for graph neural learning.


\subsection{Spatial Graph Modeling}\label{AA}
To learn discriminative spatial interaction patterns among ROIs, we employ a multi layer Graph Attention Network (GAT). Each ROI node is represented by a 64 dimensional embedding. Three GAT layers with hidden dimensions 128, 128, and 64 are then applied to model spatial dependencies among brain regions.
Given node features $\mathbf{h}_i$ and adjacency $\mathbf{A}^{(t)}$, attention coefficients between connected ROIs are computed as presented in Eq.~\ref{eq:r}.

\begin{equation}
e_{ij} = \mathrm{LeakyReLU}
\left(
\mathbf{a}^T
\left[
\mathbf{W}\mathbf{h}_i \Vert 
\mathbf{W}\mathbf{h}_j
\right]
\right)
\label{eq:r}
\end{equation}

Normalized attention weights are obtained via the softmax operation in Eq.~\ref{eq:rw}.

\begin{equation}
\alpha_{ij} = 
\frac{\exp(e_{ij})}
{\sum_{k \in \mathcal{N}(i)} \exp(e_{ik})}
\label{eq:rw}
\end{equation}

Node embeddings are updated through attention-weighted aggregation, as defined in Eq.~\ref{eq:rws}.

\begin{equation}
\mathbf{h}_i' = 
\sum_{j \in \mathcal{N}(i)} 
\alpha_{ij}\mathbf{W}\mathbf{h}_j
\label{eq:rws}
\end{equation}

Finally, an attention-based pooling module ($64\rightarrow32 \rightarrow 1$) aggregates node embeddings into a compact spatial feature vector $\mathbf{z}^{(t)}$ for each temporal window.
\subsection{Temporal Dynamics Modeling}\label{AA}
Embeddings from all windows are arranged as a temporal sequence $\mathbf{Z}$, as expressed in Eq.~\ref{eq:rs}.

\begin{equation}
\mathbf{Z} = 
[\mathbf{z}^{(1)}, \mathbf{z}^{(2)}, \dots, \mathbf{z}^{(T)}]
\label{eq:rs}
\end{equation}

Temporal dynamics are captured using a 1D convolution layer with 96 filters and kernel size 3, followed by batch normalization and ReLU activation. The convolutional operation is formulated in Eq.~\ref{eq:ws}.

\begin{equation}
\mathbf{u} = \mathrm{Conv1D}(\mathbf{Z})
\label{eq:ws}
\end{equation}

To emphasize the most discriminative temporal segments, a temporal attention network ($96 \rightarrow 48 \rightarrow 1$) assigns importance weights, as computed in Eq.~\ref{e:ws}.

\begin{equation}
\beta_t = 
\frac{\exp(g(\mathbf{u}_t))}
{\sum_{k=1}^{T}\exp(g(\mathbf{u}_k))}
\label{e:ws}
\end{equation}

The final temporal feature vector is computed as a weighted sum of window features, as defined in Eq.~\ref{e:wsb}.

\begin{equation}
\mathbf{v} = \sum_{t=1}^{T}\beta_t\mathbf{u}_t
\label{e:wsb}
\end{equation}


\subsection{Classification Module}\label{AA}
The aggregated feature representation $\mathbf{v}$ is passed through two fully connected layers with 64 and 32 units with dropout regularization, followed by a softmax function. The final prediction is obtained as presented in Eq.~\ref{eq:wsb}.

\begin{equation}
\hat{y} = \mathrm{Softmax}(f(\mathbf{v}))
\label{eq:wsb}
\end{equation}

where $\hat{y}$ denotes the predicted probability of ADHD versus healthy control (HC). 

\subsection{Attention-Based Interpretability} 
To enhance clinical transparency, the proposed framework provides interpretable attention outputs. Graph attention weights $\alpha_{ij}$ highlight dominant ROI-to-ROI connectivity patterns within the functional brain network, offering insight into influential spatial interactions. In addition, ROI importance scores are derived to identify brain regions contributing most significantly to the final prediction. Temporal attention weights $\beta_t$ are incorporated to emphasize informative dynamic connectivity segments within the model. Collectively, these outputs facilitate the identification of potential neuroimaging biomarkers.

\section{Experimental Evaluation}
\subsection{Dataset Acquisition and Preprocessing}\label{AA}
Resting-state fMRI (rs-fMRI) scans have been obtained from the Peking University (PU) site of the publicly available ADHD-200 dataset, which includes 194 subjects (116 HCs and 78 ADHD patients). These rs-fMRI scans capture spontaneous functional brain activity, with each scan having a spatial resolution of  $49 \times 58 \times 47$ voxels over $232$ time points. All data have been preprocessed using the Athena pipeline ~\cite{b19}. An ROI-based analysis is adopted, where the brain is  parcellated using the Automated Anatomical Labeling (AAL) atlas ~\cite{b20} into 116 regions, enabling extraction of regional time series for subsequent functional connectivity modeling.
\subsection{Classification Performance}\label{AA}
\begin{figure}[t]
\centering
\includegraphics[width=0.8\linewidth]{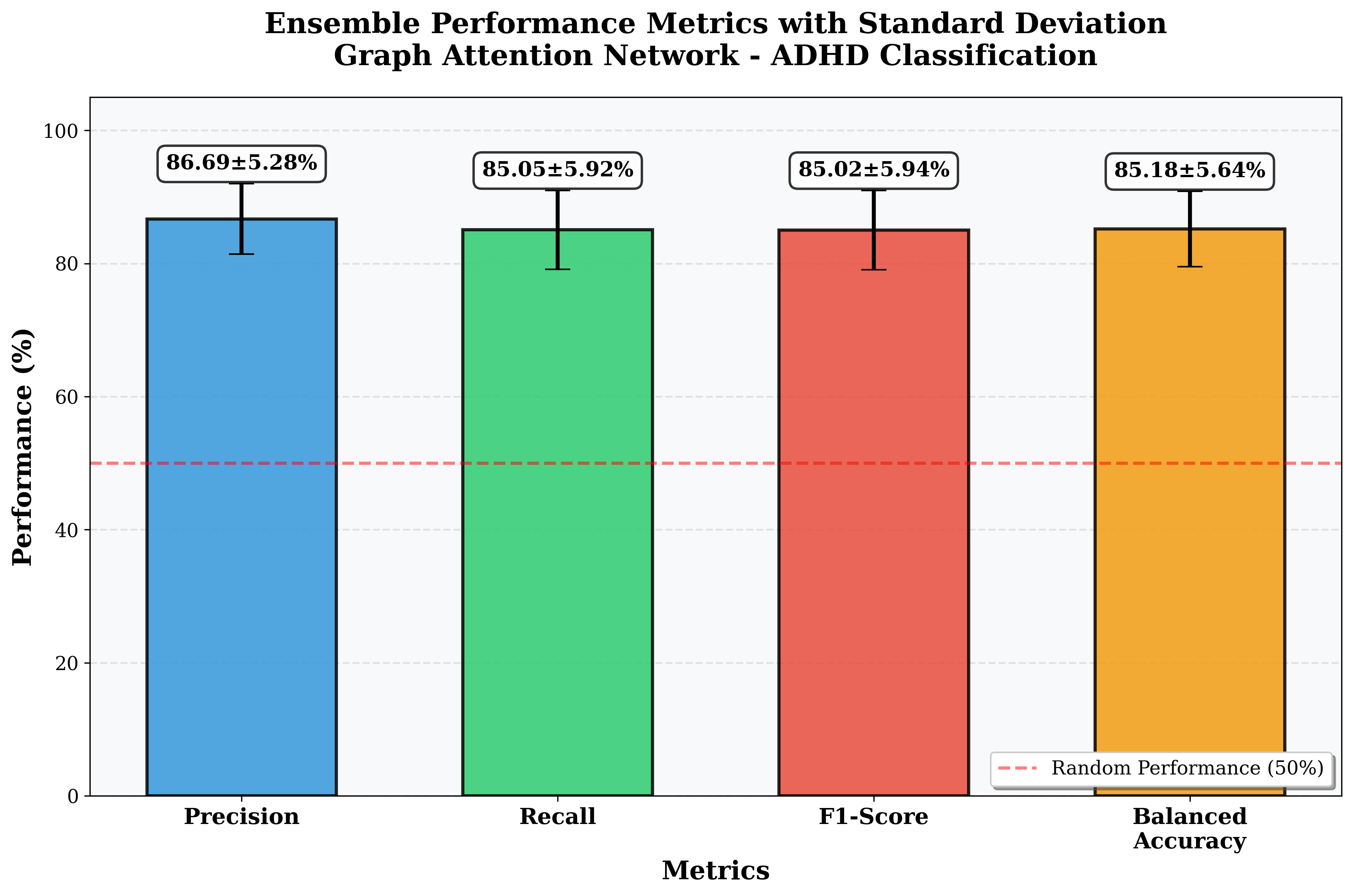}
\caption{Performance of the proposed D-GATNet framework for ADHD classification. Error bars indicate mean $\pm$ standard deviation across folds.}
\label{fig:performance}
\end{figure}

Dynamic FC matrices are generated using a sliding-window scheme, where the window length and step size are empirically selected by testing multiple configurations. The best performance is obtained with a window size of 40 time points and a step size of 20. Model evaluation is performed using stratified 10 fold cross-validation to ensure balanced subject distribution across folds. To improve robustness, accuracy, and reproducibility, a seed-based ensemble strategy is adopted, where the same architecture is trained with different random seeds~\cite{21}. Specifically, a 5-seed ensemble is employed within each fold, and final predictions are obtained via majority voting. The network is optimized end to end using AdamW with a learning rate of 0.001 and weight decay of $1\times10^{-4}$. Training is conducted with a batch size of 16 for up to 300 epochs, with early stopping applied (patience = 100) after a minimum of 40 epochs based on validation balanced accuracy. To mitigate overfitting, dropout regularization (0.5 and 0.4) is incorporated in the classification layers, and weighted cross-entropy loss is used to address class imbalance. Performance is assessed using balanced accuracy (ACC), precision (PRE), recall (REC), F1-score (F1), and AUC. Balanced accuracy is emphasized due to class imbalance in the dataset. Across folds, the proposed framework achieves a mean balanced accuracy of $85.18\% \pm 5.64$ and an AUC of 0.881. As shown in Fig. ~\ref{fig:performance} the model further reports precision, recall, and F1-score of $86.69\% \pm 5.28$, $85.05\% \pm 5.92$, and $85.02\% \pm 5.94$, respectively, demonstrating reliable discrimination between ADHD and HCs.

\subsection{Interpretability Results }\label{AA}

\begin{figure}[t]
\centering
\includegraphics[width=0.60\linewidth]{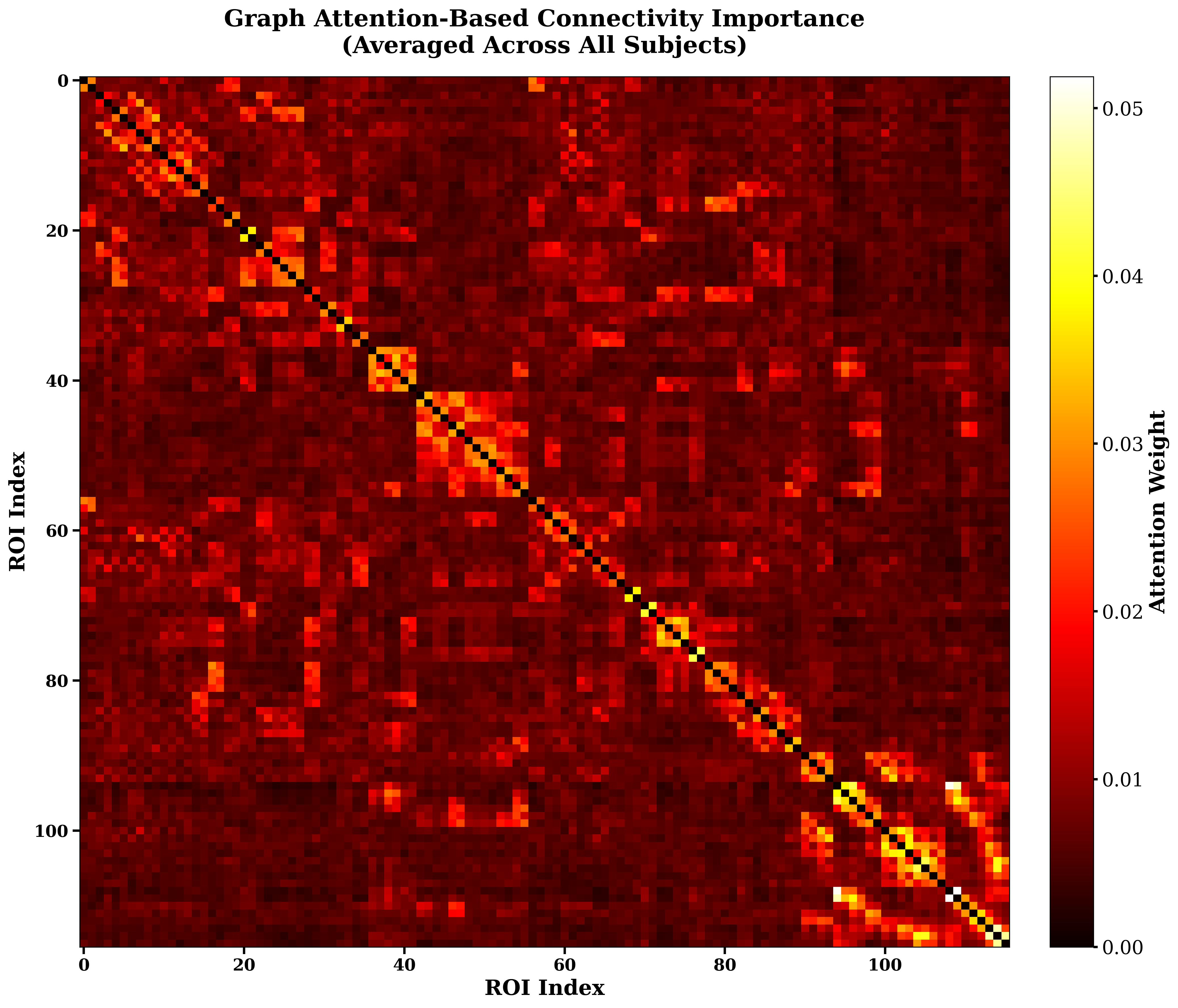}
\caption{Graph attention heatmap highlighting dominant ROI-to-ROI connectivity patterns.}
\label{fig:gat_heatmap}
\end{figure}

\begin{figure}[t]
\centering
\includegraphics[width=0.85\linewidth]{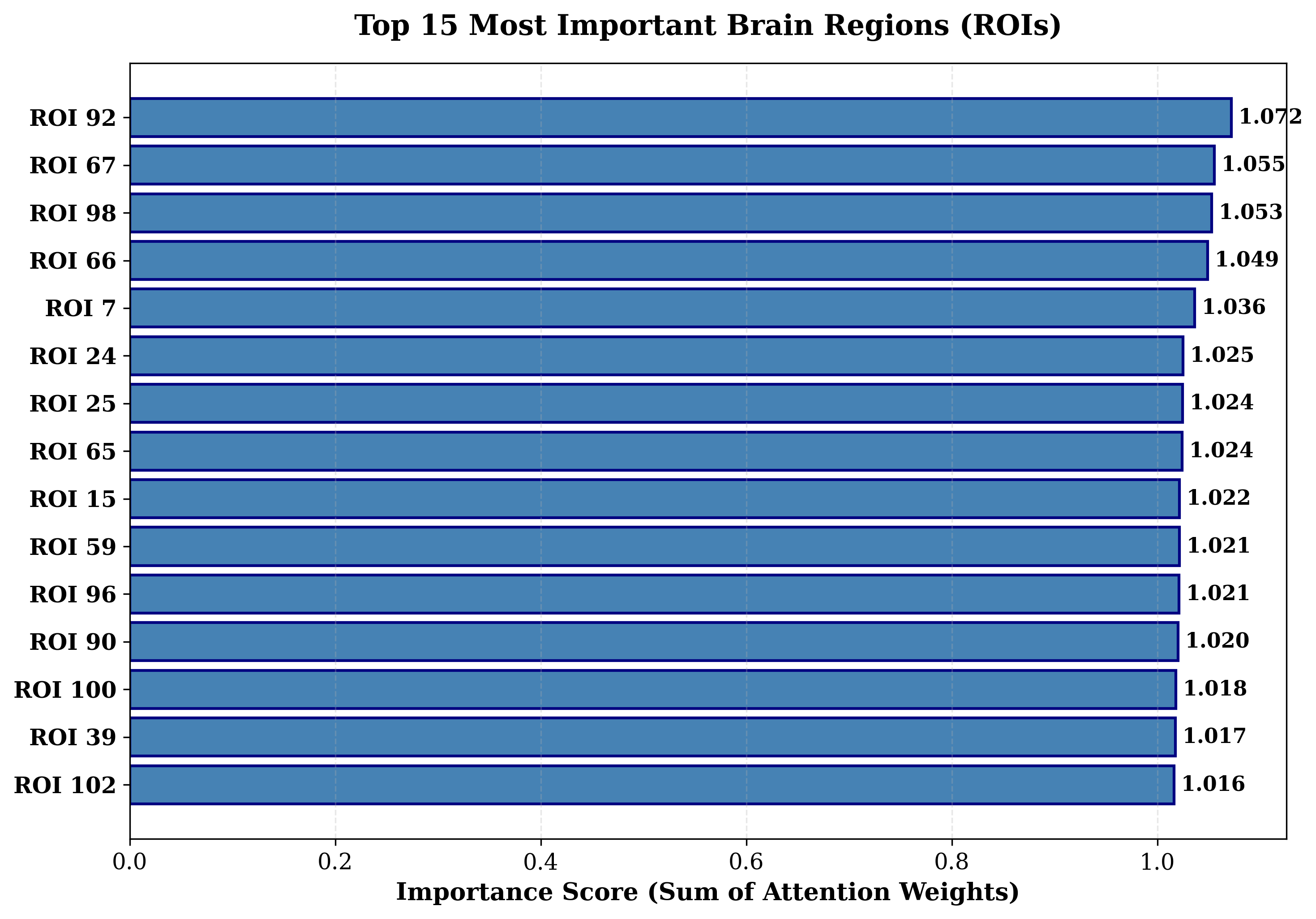}
\caption{ROI importance scores indicating the most discriminative brain regions for ADHD classification.}
\label{fig:roi_importance}
\end{figure}

To improve clinical transparency, we analyzed the learned spatial and temporal attention weights of the proposed temporal graph attention framework. 
The graph attention map highlights discriminative ROI-to-ROI connections, with the strongest interactions primarily observed among cerebellar regions and parietal hubs, as illustrated in Fig.~\ref{fig:gat_heatmap}. 
In particular, the most influential ROIs included the right Cerebellum Crus I (ROI 92), left Precuneus (ROI 67), right Angular gyrus (ROI 66), and medial frontal regions (ROI 7 and ROI 24). These areas align with well-established ADHD neurobiology: cerebellar alterations, especially within Crus I, have been linked to deficits in attentional and cognitive modulation; medial frontal abnormalities are associated with impaired executive control and response inhibition; and the precuneus and angular gyrus, key hubs of the default mode and fronto-parietal networks, have been repeatedly implicated in disrupted self-referential processing and attentional regulation in ADHD ~\cite{b22}~\cite{b23}.
Furthermore, ROI importance scores (Fig.~\ref{fig:roi_importance}) further confirm the contribution of these regions to the final prediction. 
Overall, these attention derived biomarkers provide interpretable evidence of spatio-temporal connectivity disruptions associated with ADHD.

\subsection{Ablation Study}\label{AA}
Table~\ref{tab:ablation} demonstrates the impact of major components in D-GATNet. Replacing dFC with static FC reduces performance, highlighting the importance of temporal connectivity dynamics. Removing the seed ensemble further degrades accuracy and stability, confirming its role in robust prediction. Overall, the full D-GATNet achieves the best results.
\begin{table}[h!]
\centering
\caption{Ablation study of the D-GATNet  showing classification performance.}
\label{tab:ablation}
\begin{tabular}{|p{2cm}|p{1.2cm}|p{1.2cm}|p{1.2cm}|p{1.2cm}|}
\hline
\textbf{Pipeline Variant} & \textbf{ACC (\%)} & \textbf{PRE (\%)} & \textbf{REC (\%)} & \textbf{F1 (\%)} \\
\hline
Static FC instead of dFC & 80.08 $\pm$  5.61 &  82.61 $\pm$ 5.06
 & 80.87 $\pm$ 4.30 & 80.58 $\pm$ 4.40 \\
\hline

Without Seed Ensemble & 77.10 $\pm$ 7.27 & 79.44 $\pm$ 7.34
 & 77.18 $\pm$ 7.35& 77.01 $\pm$ 7.28 \\
\hline
D-GATNet  & 85.18 $\pm$  5.64 & 86.69 $\pm$ 5.28, & 85.05 $\pm$ 5.92 & 85.02 $\pm$ 5.94 \\
\hline
\end{tabular}
\end{table}

\subsection{Performance Comparison}\label{AA}
The performance of the proposed approach is compared with several state of the art methods using rs-fMRI data from the Peking University site of the ADHD-200 dataset for fair evaluation. As shown in Table~~\ref{tab:comparison_sota}, Qiang et al.~\cite{b8} employed a spatiotemporal attention autoencoder with 5-fold CV, achieving 79.5\% accuracy but without interpretability. Wang et al. ~\cite{b11} utilized ICA-based connectivity features with CNN and autoencoder models, reporting performance 67\%. Zhang et al. ~\cite{b13} introduced a diffusion kernel attention network on functional brain networks, providing interpretability but reaching only 70.6\% accuracy on PU data. Similarly, Qiu et al.~\cite{b17} applied dFC features with ASTNet, obtaining 74.5\%, while Gulhan and Özmen ~\cite{b15} used faLFF and ReHo features with a 3D-CNN, achieving 70.37\%. In contrast, the proposed D-GATNet leverages dynamic functional connectivity with a temporal multi-head graph attention framework, outperforming prior studies with an accuracy of 85.18\% while also offering interpretable attention-based potential biomarkers.

\begin{table}[t]
\centering
\caption{Comparison of the proposed D-GATNet with state-of-the-art rs-fMRI-based ADHD classification methods on the Peking University site of the ADHD-200 dataset.}
\label{tab:comparison_sota}
\renewcommand{\arraystretch}{1.1}
\setlength{\tabcolsep}{3pt}

\begin{tabular}{|p{1.8cm}|p{1.3cm}|p{1.2cm}|p{1.3cm}|p{1.2cm}|p{1.0cm}|}
\hline
\textbf{Method} & \textbf{Features} & \textbf{Model}   & \textbf{Validation} & \textbf{ACC.}& \textbf{Interp.} \\
\hline
Qiang et al.[8] & Spatio-temporal & STAAE   & 5-fold CV& 79.5\%& No \\
\hline
Wang et al.[11]& ICA, connectivity & ICA-CNN, Corr-AE   & Holdout & 67.0\%& No \\
\hline
Zhang et al.[13] & FBN connectivity & Diffusion Kernel Attn. Net   & Holdout& 70.6\%& Yes \\
\hline
Qiu et al.[17] & dFC & ASTNet  & Holdout& 74.5\% & No\\
\hline
Gulhan \& Özmen [15] & faLFF, ReHo & 3D-CNN   & 5-fold CV & 70.37\%& No \\
\hline
\textbf{D-GATNet (Ours)} & \textbf{dFC} & \textbf{Temporal GAT-GNN}   & \textbf{Stratified 10-fold CV}& \textbf{85.18\%}& \textbf{Yes} \\
\hline
\end{tabular}
\end{table}

\section{Conclusion and Future Work}

In this study, we introduced D-GATNet, a dynamic graph attention framework for ADHD identification from rs-fMRI. By leveraging dynamic functional connectivity, the proposed method effectively captures time-varying brain network alterations through spatial graph attention and temporal modeling. The framework achieved strong classification performance on the ADHD-200 Peking University dataset, surpassing several existing deep learning approaches. Importantly, D-GATNet enhances clinical transparency by providing attention-driven biomarkers that highlight influential ROI interactions, particularly involving cerebellar and default-mode network regions. These results demonstrate the potential of interpretable temporal graph learning for robust neuroimaging-based ADHD diagnosis. Future research will focus on extending D-GATNet toward multi-site generalization and larger-scale evaluation to improve clinical applicability. Incorporating multimodal neuroimaging data and adaptive windowing strategies may further enhance connectivity modeling.

\end{document}